\documentclass[letterpaper]{article} 
\usepackage[preprint]{aaai2027}  
\usepackage[hyphens]{url}  
\usepackage{graphicx} 
\usepackage{natbib}  
\usepackage{caption} 
\usepackage{algorithm}
\usepackage{algorithmic}

\usepackage{newfloat}
\usepackage{listings}
\DeclareCaptionStyle{ruled}{labelfont=normalfont,labelsep=colon,strut=off} 
\floatstyle{ruled}
\newfloat{listing}{tb}{lst}{}
\floatname{listing}{Listing}

\usepackage{booktabs}
\usepackage{amsmath,amssymb}
\usepackage{bm}

\title{The Visual Target Matters: Learning across the Visual Hierarchy for Brain-to-Image Retrieval}
\author{
Ye Wang\textsuperscript{\rm 1},
Haokun Ren\textsuperscript{\rm 1},
Hong Yu\textsuperscript{\rm 1},
Ruirui Li\textsuperscript{\rm 3},
Xiao Li\textsuperscript{\rm 4},
Ke Liu\textsuperscript{\rm 1}\corresponding,
Wei Wu\textsuperscript{\rm 2}
}

\affiliations{
\textsuperscript{\rm 1}School of Artificial Intelligence, Chongqing University of Posts and Telecommunications, Chongqing, China\\
\textsuperscript{\rm 2}School of Medicine, Shanghai Jiaotong University, Shanghai, China\\
\textsuperscript{\rm 3}College of Information Science and Technology, Beijing University of Chemical Technology, Beijing, China\\
\textsuperscript{\rm 4}Department of Land Surveying and Geo-Informatics, The Hong Kong Polytechnic University, Hong Kong, China\\
liuke@cqupt.edu.cn
}

\begin{document}

\maketitle

\begin{abstract}
Brain-to-image retrieval seeks to identify the visual stimulus that elicited a non-invasive neural response. Candidate images are typically represented by pretrained vision models, whose internal representations vary in abstraction across depth. Existing methods usually train the neural encoder to recover a fixed final-layer visual target. Under this formulation, the visual hierarchy is reduced to a single prescribed endpoint, preventing representations at other depths from directly shaping the visual target. This limitation motivates learning how information across visual depths should contribute to the retrieval target. To this end, we introduce \textbf{NeuroGlyph}, which learns a trial-independent visual target from multiple depths of a frozen visual backbone. NeuroGlyph decomposes the target into factor-specific subspaces. Each subspace learns an image-conditioned allocation over visual depth. The resulting subspaces are fused into a single embedding for retrieval. Across THINGS-EEG and THINGS-MEG, NeuroGlyph outperforms final-layer supervision in all controlled comparisons. It also surpasses the post hoc best fixed-layer oracle in three of four comparisons. Parameter-matched ablations support both factorized target construction and image-conditioned depth allocation. Under comparable 200-way retrieval protocols, NeuroGlyph achieves the strongest system-level performance in six of eight reported metrics. These results support learning retrieval targets across the visual hierarchy rather than prescribing one visual depth.
\end{abstract}


\section{Introduction}
\label{sec:introduction}

Brain-to-image retrieval seeks to identify the visual stimulus that elicited a non-invasive neural response \cite{zhang2025cognitioncapturer}. A neural encoder is typically trained to align an electroencephalography (EEG) or magnetoencephalography (MEG) recording with the representation of its corresponding image \cite{wu2025ubp}. Most existing methods use the final-layer output of a pretrained vision model as the prescribed visual target \cite{du2023bravl}. With this target fixed in advance, subsequent research has primarily emphasized improving how neural measurements are mapped to that representation \cite{song2024nice}.

Prior studies have shown that pretrained vision models form a hierarchical representation space, in which different depths capture distinct combinations of appearance, spatial structure, object parts, and semantic content \cite{li2024guided,zheng2026hierarchical,zhu2026rahs}. However, this hierarchy is largely overlooked when the final-layer representation is prescribed as the sole visual target. Neural responses have also been associated with visual features across multiple depths rather than exclusively with the final layer \cite{liu2025vieeg}. Therefore, restricting supervision to the final representation excludes visual information that is more compatible with the available neural signals.

\begin{figure}[t!]
    \centering
    \includegraphics[width=1.00\linewidth]{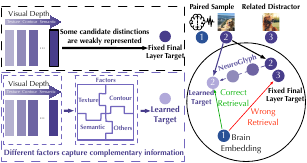}
    \caption{
\textbf{The visual target shapes brain-to-image retrieval.}
Final-layer supervision poorly distinguishes a paired image from related distractors. NeuroGlyph constructs a unified target with complementary evidence across visual depths through factor-specific, image-conditioned allocation.
}
    \label{fig:motivation}
\end{figure}

An intermediate layer provides a different prescribed target, yet the underlying fixed-layer formulation remains unchanged \cite{du2026deep}. As illustrated in Figure~\ref{fig:motivation}, this formulation treats the visual hierarchy as a set of alternative endpoints, from which one layer must be selected in advance. It cannot determine how information across visual depths should jointly shape the retrieval target. Therefore, the central problem is to learn the target from the visual hierarchy rather than prescribe a single visual depth.

Consequently, a target derived from the visual hierarchy must account for variation across both target components and images. A single depth weighting forces all components to share the same abstraction profile, whereas a uniform weighting across images ignores differences in the visual evidence needed for discrimination. The target constructor should combine complementary information across depths and adapt their contributions to the target component and image content. Each gallery image must also retain a fixed target independent of the neural trial used as the query.

To this end, we introduce \textbf{NeuroGlyph}, which learns a trial-independent visual target from multiple depths of a frozen visual backbone. NeuroGlyph decomposes the target into factor-specific subspaces. Each subspace learns an image-conditioned allocation over visual depth. The resulting subspaces are fused into a
single embedding for retrieval. Across THINGS-EEG and THINGS-MEG, NeuroGlyph outperforms final-layer supervision in all controlled comparisons and surpasses the post hoc best fixed-layer oracle in three
of four. Parameter-matched ablations support factorized target construction and image-conditioned depth allocation. Under comparable 200-way retrieval protocols, NeuroGlyph achieves the strongest system-level performance in six of eight reported metrics.

Our main contributions are summarized below.
\begin{itemize}
\item We formulate brain-to-image retrieval as learning visual targets from the hierarchy of a pretrained vision model, incorporating target construction into the task explicitly.

\item We propose NeuroGlyph, which constructs a trial-independent visual target by leveraging complementary information across multiple visual depths.

\item NeuroGlyph consistently outperforms final-layer supervision, surpassing the best fixed-layer target selected post hoc in three of the
four evaluation settings. Parameter-matched ablations further validate the effectiveness of the proposed target construction strategy.

\end{itemize}

\section{Related Work}
\label{sec:related_work}

\subsection{Neural--Visual Representation Alignment}

Most brain-to-image decoding methods map EEG or MEG responses into a
pretrained visual or multimodal space \cite{du2023bravl,song2024nice}. Existing approaches employ brain--vision--language representations, contrastive retrieval, multimodal supervision, diffusion priors, and semantic or cognitive guidance \cite{li2024guided,zhang2025cognitioncapturer,yu2026d2fosa}.
Other studies transfer visual priors to EEG representations by leveraging knowledge distillation, cross-domain learning, algorithm unrolling, bidirectional fusion, and multimodal disentanglement \cite{xu2025brainvision,zhou2026eegit,qu2026neurovision,xu2026neurovista}.
Related principles have also been extended to 3D visual decoding \cite{guo2025neuro3d}.

These methods improve neural encoding and cross-modal alignment, but
their supervision generally remains anchored to representations
produced by pretrained visual models. They focus primarily
on mapping neural responses into a given visual space rather than
constructing the retrieval target from its internal hierarchy.

\subsection{Adaptive Visual Supervision}

Recent methods adapt visual supervision through blur or linguistic
priors, semantic--perceptual interpolation and teacher
compression, or learned visual-space adaptation \cite{wu2025ubp,wu2025shrinking,liu2026blur,liu2026linguistic,jo2026hyfi}. Specifically, NeuroBridge combines modality-specific augmentation, multi-view visual aggregation, and shared semantic projection, while CFT-NET progressively adapts visual and semantic representations \cite{zhang2026neurobridge,sun2026crossmodal}.
Brain-aligned semantic vectors, human-aligned encoders, and ReAlnet
further reshape visual supervision toward neural or human representational geometry \cite{rajabi2025humanaligned,vafaei2026brainaligned,lu2026realnet}.

Complementary neural-side works improve generalization through
multi-subject modeling, large-scale EEG or EEG--MEG pretraining, and language-model-guided EEG representation learning \cite{liu2024llmstates,wang2024eegpt,wang2025zebra,barmpas2025labrampp,xiao2025brainomni,liu2026mindcross}.
Together, these methods show that visual and neural representations
can be adapted, but they typically modify inputs, teachers, endpoints,
or global alignment spaces rather than factorizing information across
visual depths.

\subsection{Hierarchical and Intermediate-Layer Targets}

Hierarchical methods use biologically motivated streams, regional
features, hierarchical CLIP representations, multi-level
cross-attention, or multiple pretrained encoders \cite{liu2025vieeg,zhu2026rahs,lee2026definealignfuse,
yao2026brainssd,zheng2026hierarchical}.
They demonstrate the value of complementary visual information, but
usually organize it through predefined streams, heterogeneous sources,
or a unified fusion representation.

The most closely related model, Shallow Alignment, shows that an
intermediate visual layer can be more compatible with neural
measurements than the final layer \cite{du2026deep}. However, selecting a single layer still yields a single global target and does not exploit complementary evidence across different depths. Conventional multi-layer methods likewise combine layers via averaging, concatenation, scalar weighting, or shared fusion, typically producing a single globally fused target.

Overall, existing methods still prescribe visual targets, leaving cross-depth contributions unresolved. A more flexible target construction should combine complementary depths and adjust their contributions across representation components and images.

\section{Preliminaries}
\label{sec:preliminaries}

\subsection{Brain-to-Image Retrieval with Fixed Visual Targets}
\label{sec:fixed_target_retrieval}

Consider paired neural recordings and visual stimuli
$\mathcal{D}=\{(\bm{b}_i,\bm{x}_i)\}_{i=1}^{N}$, where $\bm{b}_i$ is
an EEG or MEG recording elicited by image $\bm{x}_i$. A neural encoder
maps the recording to a normalized query embedding
\begin{equation}
    \bm{e}_i
    =
    \operatorname{norm}\!\left(
        E_{\theta}(\bm{b}_i)
    \right)
    \in\mathbb{R}^{d},
    \label{eq:prelim_brain_embedding}
\end{equation}
where $\operatorname{norm}(\bm{a})=\bm{a}/\|\bm{a}\|_2$. A standard
fixed target uses the final image-level representation
$\bm{c}_f(\bm{x}_i)\in\mathbb{R}^{d_c}$ from a pretrained visual
backbone:
\begin{equation}
    \bm{y}^{\mathrm{fixed}}_i
    =
    \operatorname{norm}\!\left(
        \bm{P}_f\bm{c}_f(\bm{x}_i)
    \right)
    \in\mathbb{R}^{d},
    \label{eq:fixed_final_target}
\end{equation}
where $\bm{P}_f\in\mathbb{R}^{d\times d_c}$ maps the final feature to
the retrieval dimension. The retrieval score between neural query $i$
and candidate image $j$ is
\begin{equation}
    s_{ij}
    =
    \bm{e}_i^{\top}\bm{y}^{\mathrm{fixed}}_j.
    \label{eq:fixed_retrieval_score}
\end{equation}
The query encoder is learned, but the representational content of the
visual endpoint is prescribed before neural--visual alignment.

\subsection{Visual-Target Supervision Bottleneck}
\label{sec:target_bottleneck}

For the following argument, let $\bm{y}_i$ denote an arbitrary
normalized visual target assigned to image $\bm{x}_i$, including but not
limited to the fixed final-layer target in
Equation~\eqref{eq:fixed_final_target}. For the correct image
$\bm{x}_i$ and an incorrect candidate $\bm{x}_j$, define the pairwise
retrieval margin as
\begin{equation}
\begin{aligned}
    m_{i\rightarrow j}
    &=
    s_{ii}-s_{ij} \\
    &=
    \bm{e}_i^{\top}
    \left(
        \bm{y}_i-\bm{y}_j
    \right).
\end{aligned}
\label{eq:retrieval_margin}
\end{equation}
For normalized neural embeddings, the Cauchy--Schwarz inequality gives
\begin{equation}
    \left|m_{i\rightarrow j}\right|
    \leq
    \left\|\bm{y}_i-\bm{y}_j\right\|_2.
    \label{eq:retrieval_margin_bound}
\end{equation}
Thus, visual-target geometry constrains the candidate distinctions that
can be expressed under the single-embedding cosine scoring rule in
Equation~\eqref{eq:fixed_retrieval_score}. If two images receive
identical targets, no neural encoder can separate them under this rule;
if their targets are close, the attainable pairwise margin is
correspondingly limited. The visual target is therefore not a passive
coordinate system.

This bound alone does not determine target quality. An effective target must balance \emph{candidate discriminability} with \emph{neural predictability}. Final-layer representations may separate images well while suppressing perceptual or structural cues recoverable from EEG or MEG. We term this mismatch the \emph{visual-target supervision bottleneck}: the final layer remains informative but may not be the most compatible neural retrieval target.

\subsection{From a Fixed Target to a Learnable Target}
\label{sec:learnable_target}

We replace the prescribed endpoint with a target constructed from a set
of internal visual representations:
\begin{equation}
    \bm{z}_{\phi}(\bm{x})
    =
    \operatorname{norm}\!\left(
        F_{\phi}\!\left(
            \{\bm{h}_{\ell}(\bm{x})\}_{\ell\in\mathcal{S}};
            \bm{c}_f(\bm{x})
        \right)
    \right),
    \label{eq:conceptual_target_constructor}
\end{equation}
where $\mathcal{S}$ denotes selected visual depths and $F_{\phi}$ is a
learnable image-side target constructor. The neural encoder and target
constructor are jointly optimized by a retrieval objective:
\begin{equation}
    (\theta^{\star},\phi^{\star})
    =
    \arg\min_{\theta,\phi}
    \mathcal{L}_{\mathrm{ret}}
    \left(
        \{\bm{e}_i\}_{i=1}^{N},
        \{\bm{z}_{\phi}(\bm{x}_i)\}_{i=1}^{N}
    \right).
    \label{eq:joint_target_optimization}
\end{equation}
Positive pairs impose neural-compatibility pressure, whereas negative
candidates prevent the target constructor from discarding image
information merely because it is easy to predict.

These observations yield three requirements for visual-target construction: it should draw on multiple visual depths, remain trial-independent while being shaped by neural supervision during training, and preserve complementary depth-specific evidence without a shared global preference. Next, NeuroGlyph is proposed accordingly.

\section{Method}
\label{sec:method}

\begin{figure*}[t]
    \centering
    \includegraphics[width=\textwidth]{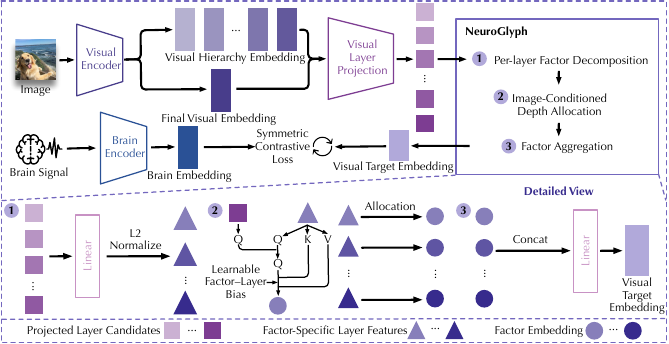}
    \caption{\textbf{Overview of NeuroGlyph.} A frozen visual backbone extracts representations across multiple depths. The factorized target constructor learns image-conditioned depth allocations for different factors and combines them into a unified visual target. The constructor and neural encoder are jointly optimized through symmetric contrastive learning. Gallery targets depend only on image features and can be precomputed for retrieval.}
    \label{fig:method_overview}
\end{figure*}

\subsection{Overview}
\label{sec:method_overview}

Given $\mathcal{D}=\{(\bm{b}_i,\bm{x}_i)\}_{i=1}^{N}$, a frozen visual
backbone extracts token-level representations
$\bm{H}_{i\ell}=\bm{h}_{\ell}(\bm{x}_i)$ at selected depths
$\ell\in\mathcal{S}$ and a final image-level feature
$\bm{c}_i=\bm{c}_f(\bm{x}_i)$. NeuroGlyph uses a learnable target
constructor to map these image features to a normalized target
$\bm{z}_i\in\mathbb{R}^{d}$. In parallel, a learnable neural encoder
maps $\bm{b}_i$ to a normalized query $\bm{e}_i\in\mathbb{R}^{d}$.
Only the neural encoder and target constructor are updated; the visual
backbone remains frozen.

\subsection{Hierarchical Visual Candidates}
\label{sec:hierarchical_candidates}

For each selected depth $\ell\in\mathcal{S}$, we pool the patch tokens
and project them into a shared $d$-dimensional space:
\begin{equation}
    \bm{v}_{i\ell}
    =
    \operatorname{norm}\!\left(
        \bm{P}_{\ell}
        \operatorname{Pool}(\bm{H}_{i\ell})
    \right),
    \label{eq:layer_projection}
\end{equation}
where $\bm{P}_{\ell}$ is specific to visual depth $\ell$. Independent
projections account for depth-dependent feature statistics while making
the selected representations comparable within the target constructor.
We use mean pooling over patch tokens unless stated otherwise.

\subsection{Factorized Visual Target Construction}
\label{sec:factorized_target}

Let $L=|\mathcal{S}|$ denote the number of selected visual depths and $K$ the number of learnable target subspaces, termed factors. These factors have no predefined semantic or biological meanings. A monolithic
depth-weighting constructor represents the entire target using one
allocation in the simplex $\Delta^{L-1}$. NeuroGlyph instead learns
\begin{equation}
    \left(
        \bm{\alpha}_1(\bm{x}),\ldots,
        \bm{\alpha}_K(\bm{x})
    \right)
    \in
    \left(\Delta^{L-1}\right)^K,
    \label{eq:factorized_allocation_space}
\end{equation}
where
$\Delta^{L-1}=\{\bm{a}\in\mathbb{R}_{\geq0}^{L}:
\sum_{\ell=1}^{L}a_{\ell}=1\}$. Enforcing
$\bm{\alpha}_1=\cdots=\bm{\alpha}_K$ yields a shared-depth-preference
variant, whereas independent allocations allow separate target
subspaces to read the hierarchy differently.

\subsubsection{Per-Layer Factor Decomposition}

Each projected layer feature is transformed into $K$ factor-specific
channels of dimension $d_f$:
\begin{equation}
\begin{gathered}
    \begin{bmatrix}
        \widetilde{\bm{u}}_{i\ell1}^{\top}\\
        \vdots\\
        \widetilde{\bm{u}}_{i\ell K}^{\top}
    \end{bmatrix}
    =
    \operatorname{reshape}_{K\times d_f}\!\left(
        \bm{A}_{\ell}\bm{v}_{i\ell}
    \right),
    \\[3pt]
    \bm{u}_{i\ell k}
    =
    \operatorname{norm}\!\left(
        \widetilde{\bm{u}}_{i\ell k}
    \right),
    \qquad k=1,\ldots,K.
\end{gathered}
\label{eq:factor_decomposition}
\end{equation}
Here, $\bm{A}_{\ell}\in\mathbb{R}^{Kd_f\times d}$ is independent for
each visual depth. Consequently, the same depth can provide different
content to different factors instead of sharing an identical value
representation across all allocations.

\subsubsection{Image-Conditioned Depth Allocation}

Each factor has a learnable base query
$\bm{q}^{(0)}_k\in\mathbb{R}^{d_f}$, conditioned on the final
image-level feature:
\begin{equation}
\begin{gathered}
    \bm{q}_{ik}
    =
    \bm{q}^{(0)}_k+\bm{R}_k\bm{c}_i,
    \qquad
    \overline{\bm{u}}_{i\ell k}
    =
    \operatorname{LN}_k(\bm{u}_{i\ell k}),
    \\[3pt]
    \bm{k}_{i\ell k}
    =
    \bm{W}^{\mathrm{key}}_k\overline{\bm{u}}_{i\ell k},
    \qquad
    \bm{w}_{i\ell k}
    =
    \bm{W}^{\mathrm{value}}_k\overline{\bm{u}}_{i\ell k}.
\end{gathered}
\label{eq:factor_qkv}
\end{equation}
Here, $\bm{R}_k\in\mathbb{R}^{d_f\times d_c}$ and
$\bm{W}^{\mathrm{key}}_k,\bm{W}^{\mathrm{value}}_k
\in\mathbb{R}^{d_f\times d_f}$. The final-layer feature serves as a
global conditioning signal rather than the prescribed supervision
target: it controls how the hierarchy is read but is not concatenated
directly into the output target.

Factor $k$ assigns weight to visual depth $\ell$ as
\begin{equation}
    \alpha_{ik\ell}
    =
    \frac{
        \exp\!\left(
            \bm{q}_{ik}^{\top}\bm{k}_{i\ell k}/\sqrt{d_f}
            +\beta_{k\ell}
        \right)
    }{
        \sum_{r\in\mathcal{S}}
        \exp\!\left(
            \bm{q}_{ik}^{\top}\bm{k}_{irk}/\sqrt{d_f}
            +\beta_{kr}
        \right)
    },
    \label{eq:factor_attention}
\end{equation}
where $\beta_{k\ell}$ is a learnable factor--layer bias. The base query
and bias encode an initial factor-specific depth preference, while the
conditioning term allows that preference to vary across images.

\subsubsection{Factor Aggregation}

Each factor independently aggregates the hierarchy:
\begin{equation}
    \bm{f}_{ik}
    =
    \operatorname{norm}\!\left(
        \sum_{\ell\in\mathcal{S}}
        \alpha_{ik\ell}\bm{w}_{i\ell k}
    \right).
    \label{eq:factor_aggregation}
\end{equation}
The factors are concatenated and projected to one retrieval target:
\begin{equation}
    \bm{z}_i
    =
    \operatorname{norm}\!\left(
        \bm{W}_o
        [\bm{f}_{i1};\ldots;\bm{f}_{iK}]
    \right),
    \label{eq:target_fusion}
\end{equation}
where $\bm{W}_o\in\mathbb{R}^{d\times Kd_f}$. Factorization expands the target family without imposing semantic or
biological interpretations on individual factors.

\subsection{Brain--Visual Alignment}
\label{sec:brain_visual_alignment}

A learnable neural encoder $G_{\theta}$ and projection $\bm{W}_b$ map
each recording to the retrieval space:
\begin{equation}
    \bm{e}_i
    =
    \operatorname{norm}\!\left(
        \bm{W}_bG_{\theta}(\bm{b}_i)
    \right).
    \label{eq:brain_embedding}
\end{equation}
The similarity between neural query $i$ and candidate target $j$ is
\begin{equation}
    s_{ij}
    =
    \bm{e}_i^{\top}\bm{z}_j.
    \label{eq:brain_visual_score}
\end{equation}
All hierarchical visual information influences retrieval only through
$\bm{z}_j$, keeping the scoring rule identical to conventional
single-embedding retrieval.

\subsection{Joint Optimization}
\label{sec:joint_optimization}

For a mini-batch of $B$ paired examples, diagonal entries of the score
matrix correspond to positive pairs, whereas off-diagonal entries serve
as in-batch negatives. We optimize the symmetric contrastive loss
\begin{equation}
\begin{aligned}
    \mathcal{L}_{\mathrm{ret}}
    =-\frac{1}{2B}\sum_{i=1}^{B}
    \Bigg[
    &\log
    \frac{\exp(s_{ii}/\tau)}
         {\sum_{j=1}^{B}\exp(s_{ij}/\tau)}
    \\
    +{}&
    \log
    \frac{\exp(s_{ii}/\tau)}
         {\sum_{j=1}^{B}\exp(s_{ji}/\tau)}
    \Bigg],
\end{aligned}
\label{eq:retrieval_loss}
\end{equation}
where $\tau$ is the contrastive temperature. The symmetric loss
provides retrieval gradients to both the neural encoder and the visual
target constructor. The neural encoder is optimized to align each
neural representation with the target constructed for its paired
image, while the target constructor is jointly optimized with the
neural branch under paired neural--image supervision. In-batch
negatives discourage collapse to an image representation that is
identical across candidates. Importantly, neural recordings affect the
target constructor only through the contrastive loss; the target for
each image is generated solely from image features.

\subsection{Training and Inference}
\label{sec:training_inference}

The visual backbone remains frozen throughout training. The neural
encoder, layer projections, factor decomposition, factor-specific
query/key/value modules, and output projection are jointly optimized.
We initialize the factor--layer biases with depth-wise Gaussian profiles:
\begin{equation}
    \beta^{(0)}_{k\ell}
    =
    -2
    \left[
        \frac{Kp_\ell-(k-\tfrac12)(L-1)}{L}
    \right]^2,
    \label{eq:gaussian_bias_initialization}
\end{equation}
where $p_{\ell}\in\{0,\ldots,L-1\}$ indexes visual depth. These
biases remain learnable and only provide distinct initial depth
preferences; they do not assign semantic roles to factors.

At inference time, the target constructor receives only image features.
Candidate targets can therefore be computed once and stored in a target
bank. Given a neural query, retrieval requires only neural encoding and
nearest-neighbor ranking by the cosine scores in
Equation~\eqref{eq:brain_visual_score}. This property also ensures that,
within a trained model, the same image is assigned the same target
regardless of which repeated neural trial is used as the retrieval
query.

\section{Experiments and Analysis}
\label{sec:experiments}

\subsection{Experimental Setup}
\label{sec:experimental_setup}
We evaluate NeuroGlyph on THINGS-EEG and THINGS-MEG \cite{gifford2022large,hebart2022things}. THINGS-EEG contains RSVP-based recordings from 10 subjects, with 1,654 training concepts (10 images and four repetitions each) and 200 concept-disjoint test concepts (one image and 80 repetitions each). Averaging repeated responses yields 16,540 training samples and 200 test samples per subject. THINGS-MEG contains recordings from four participants, with 1,654 training concepts (12 images each) and 200 concept-disjoint test concepts (one image repeated 12 times). For both datasets, responses to the same stimulus are averaged to improve the signal-to-noise ratio. Further dataset details are provided in the supplementary material. Unless otherwise stated, NeuroGlyph uses the frozen InternViT-300M-448px visual backbone and EEGProject
as the neural encoder, with separate models trained for EEG and MEG \cite{chen2024internvl,wu2025ubp}.
Full implementation details are provided in the supplementary material.

\subsection{System-Level Comparison with Prior Methods}
\label{sec:prior_comparison}

As shown in Table~\ref{tab:eeg_meg_comparison}, we compare NeuroGlyph
with representative brain-to-image retrieval systems under comparable
200-way retrieval protocols, including BraVL \cite{du2023bravl}, NICE \cite{song2024nice}, ATM-S \cite{li2024guided}, UBP \cite{wu2025ubp}, ATS \cite{wu2025shrinking}, NeuroBridge \cite{zhang2026neurobridge}, HyFI \cite{jo2026hyfi}, BrainHiVE \cite{zheng2026hierarchical}, Visual Blur Perception \cite{liu2026blur}.

\begin{table}[t]
    \centering
    \caption{
        Brain-to-image retrieval performance on EEG and MEG.
        Results are Top-1 and Top-5 accuracy (\%). Methods differ in
        architecture and training protocol; this is therefore a
        system-level comparison rather than a controlled target
        ablation.
    }
    \label{tab:eeg_meg_comparison}

    \scriptsize
    \setlength{\tabcolsep}{2.5pt}
    \renewcommand{\arraystretch}{1.05}

    \resizebox{\columnwidth}{!}{%
    \begin{tabular}{lcccccccc}
        \toprule
        \textbf{Method}
        & \multicolumn{4}{c}{\textbf{EEG}}
        & \multicolumn{4}{c}{\textbf{MEG}} \\
        \cmidrule(lr){2-5}
        \cmidrule(lr){6-9}

        &
        \multicolumn{2}{c}{\textbf{Intra-subject}}
        &
        \multicolumn{2}{c}{\textbf{Inter-subject}}
        &
        \multicolumn{2}{c}{\textbf{Intra-subject}}
        &
        \multicolumn{2}{c}{\textbf{Inter-subject}} \\

        &
        \textbf{Top-1}
        & \textbf{Top-5}
        & \textbf{Top-1}
        & \textbf{Top-5}
        & \textbf{Top-1}
        & \textbf{Top-5}
        & \textbf{Top-1}
        & \textbf{Top-5} \\
        \midrule

        BraVL
        & 5.8 & 17.5
        & 1.8 & 7.0
        & -- & --
        & -- & -- \\

        NICE
        & 16.1 & 43.6
        & 6.2 & 21.4
        & 12.8 & 36.0
        & -- & -- \\

        NICE-SA
        & 14.7 & 41.7
        & 7.0 & 23.1
        & 12.7 & 35.0
        & -- & -- \\

        NICE-GA
        & 15.6 & 42.8
        & 5.9 & 21.6
        & 14.3 & 42.3
        & -- & -- \\

        ATM-S
        & 28.5 & 60.4
        & 11.8 & 33.7
        & -- & --
        & -- & -- \\

        UBP
        & 50.9 & 79.7
        & 12.4 & 33.4
        & 26.7 & 55.2
        & 2.2 & 10.4 \\

        ATS
        & 60.2 & 86.7
        & 14.0 & 35.8
        & 32.4 & 62.3
        & 3.4 & 11.2 \\

        NeuroBridge
        & 63.2 & 89.9
        & 19.0 & 45.9
        & 32.2 & 60.8
        & 3.4 & 12.8 \\

        HyFI
        & 68.2 & 91.9
        & 15.1 & 37.2
        & 35.8 & 64.6
        & 3.2 & 11.5 \\

        BrainHiVE
        & 75.7 & 94.6
        & 20.0 & 44.1
        & 33.7 & 60.5
        & \textbf{5.4} & 15.2 \\

        Visual Blur Perception
        & 80.0 & 96.9
        & 20.0 & 48.0
        & 44.0 & 72.0
        & 5.3 & \textbf{15.9} \\

        \textbf{NeuroGlyph (Ours)}
        & \textbf{84.8} & \textbf{98.1}
        & \textbf{22.4} & \textbf{51.2}
        & \textbf{46.0} & \textbf{73.3}
        & 4.0 & 14.8 \\

        \bottomrule
    \end{tabular}%
    }
\end{table}

NeuroGlyph achieves the highest reported accuracy in six of the eight metrics, including all EEG metrics and both MEG intra-subject metrics. The remaining two metrics correspond to MEG inter-subject retrieval.

\subsection{Generalization Across Visual Backbones}
\label{sec:backbone_generalization}

\begin{figure}[t]
    \centering
    \includegraphics[width=\linewidth]{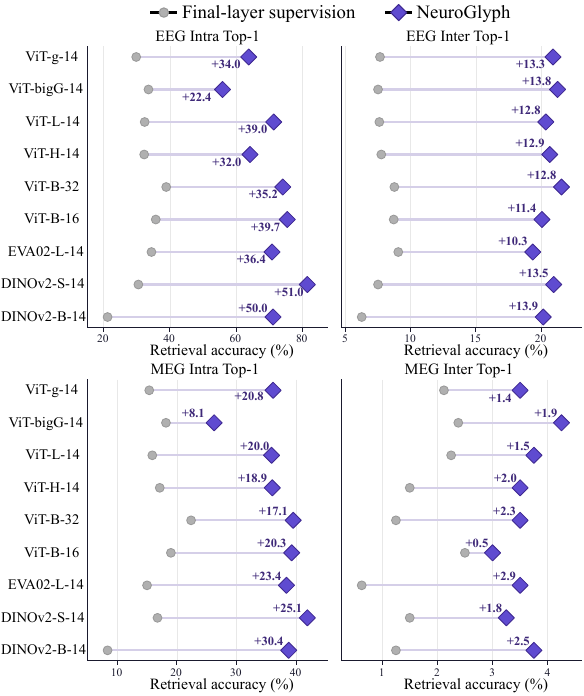}
    \caption{
        Top-1 retrieval performance across nine frozen visual
        backbones on THINGS-EEG and THINGS-MEG. Gray circles denote
        final-layer supervision, while purple diamonds denote
        NeuroGlyph. Each horizontal segment connects results obtained
        using the same visual backbone, neural encoder, and training
        protocol. Annotated values report the absolute Top-1 gain of
        NeuroGlyph in percentage points. Axes are scaled independently
        and should be compared within each panel.
    }
    \label{fig:visual_backbone_generalization}
\end{figure}

As shown in Figure~\ref{fig:visual_backbone_generalization}, we compare NeuroGlyph with final-layer supervision across nine frozen backbones spanning CLIP-style ViTs, DINOv2, and EVA-02 \cite{dosovitskiy2021vit,oquab2024dinov2,fang2024eva02}. NeuroGlyph improves all 36 backbone--setting comparisons. Top-1 gains range from 22.4--51.0 percentage points for EEG intra-subject, 8.1--30.4 for MEG intra-subject, 10.3--13.9 for EEG inter-subject, and 0.5--2.9 for MEG inter-subject retrieval. The improvement remains positive for every backbone under both neural modalities and both subject protocols, despite substantial variation in the absolute performance of the corresponding final-layer baselines.

\subsection{RQ1. Does Visual-Target Depth Matter?}
\label{sec:layerwise_compatibility}

For each visual depth, we independently train a separate model using
that depth as the sole fixed target. The neural-encoder architecture
and all other optimization and evaluation settings remain unchanged.

\begin{figure}[t]
    \centering
    \includegraphics[width=\linewidth]{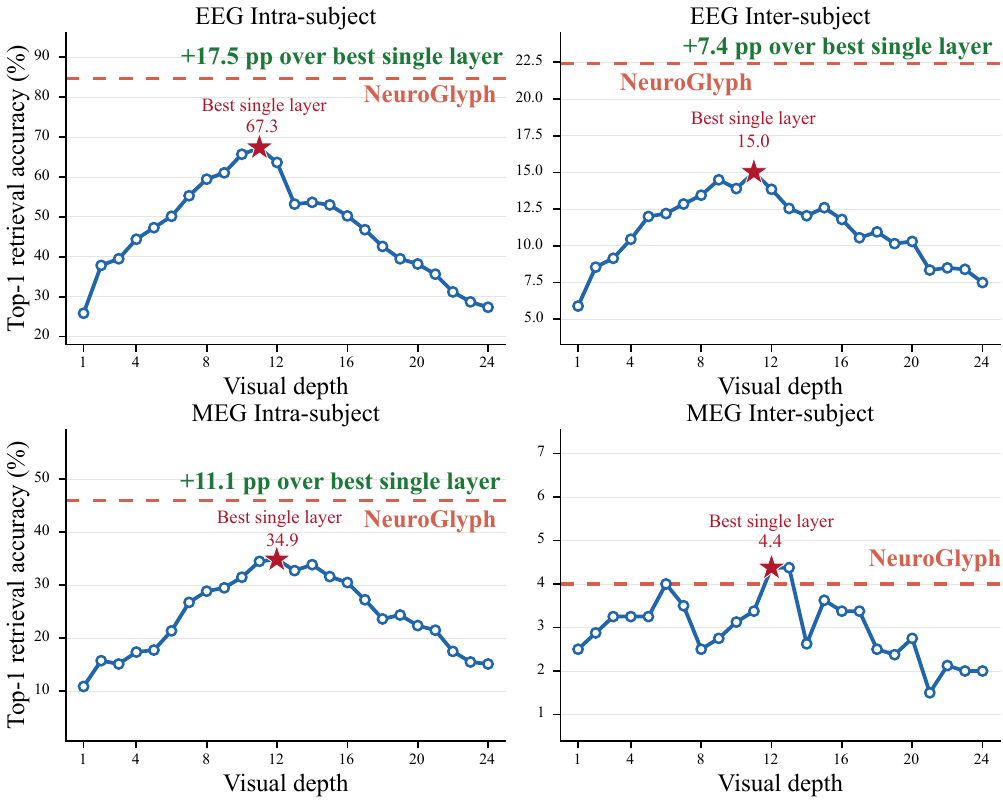}
    \caption{
        Layer-wise compatibility of fixed visual targets. Each curve reports
Top-1 accuracy obtained by independently training against one visual
depth. Red stars mark the setting-specific post-hoc best layers, and
dashed lines report NeuroGlyph.
    }
    \label{fig:layerwise_target_compatibility}
\end{figure}

As shown in Figure~\ref{fig:layerwise_target_compatibility}, retrieval
performance varies substantially with visual depth. In the
intra-subject settings, accuracy rises from shallow representations
to a broad intermediate-depth optimum and decreases again near the
visual endpoint. Inter-subject profiles are less smooth but remain
clearly depth-sensitive.

The final layer is not the strongest fixed target in any of the four
settings. However, the strongest depth also varies across neural
modalities and subject protocols. These results support two
conclusions: visual-target depth is a consequential modeling choice,
and replacing the final layer with one universally optimal
intermediate layer is not sufficient.

The setting-specific red stars provide a strong post-hoc diagnostic.
NeuroGlyph exceeds the strongest fixed layer in three of the four
Top-1 settings. In MEG inter-subject retrieval, the best fixed layer
reaches 4.4\%, compared with 4.0\% for NeuroGlyph. Thus,
NeuroGlyph is not uniformly superior to a complete setting-specific
layer search, particularly in the lowest-accuracy regime.

\subsection{RQ2. Can a Single Intermediate Layer Generalize?}
\label{sec:fixed_layer_transfer}

The previous analysis selects the best layer independently for each
evaluation setting. We now consider a stricter transfer setting. We
select one intermediate layer according to the highest EEG
intra-subject Top-1 accuracy in
Fig.~\ref{fig:layerwise_target_compatibility}, and reuse exactly the
same layer for EEG inter-subject, MEG intra-subject, and MEG
inter-subject retrieval. No layer reselection is performed for the
remaining settings.

\begin{figure}[t]
    \centering
    \includegraphics[width=\linewidth]{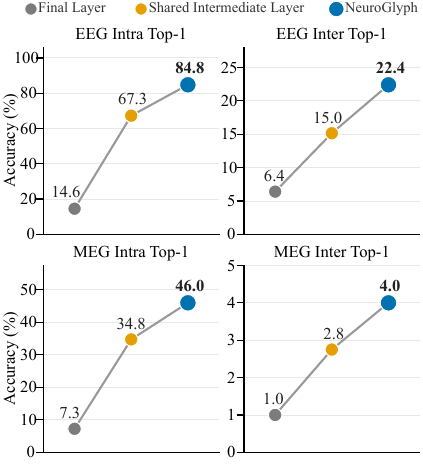}
    \caption{
        Comparison of final-layer supervision, one shared intermediate
        layer selected using EEG intra-subject performance, and
        NeuroGlyph. The intermediate layer is selected once according
        to the highest EEG intra-subject Top-1 accuracy in
        Fig.~\ref{fig:layerwise_target_compatibility}, and is then reused
        unchanged in all four settings. It is therefore different from
        the setting-specific best fixed layers marked by red stars in
        Fig.~\ref{fig:layerwise_target_compatibility}. Axes are scaled
        independently and should be compared only within each panel.
    }
    \label{fig:analysis2}
\end{figure}

As shown in Figures~\ref{fig:layerwise_target_compatibility}
and~\ref{fig:analysis2}, the shared intermediate layer consistently
outperforms final-layer supervision, including in the three settings
for which it was not selected, indicating that the advantage of an
appropriately selected intermediate representation can transfer across
neural modalities and subject protocols. NeuroGlyph further improves
over this shared layer by 17.5, 7.4, 11.2, and 1.2 percentage points
for EEG intra-subject, EEG inter-subject, MEG intra-subject, and MEG
inter-subject retrieval, respectively, suggesting that a single
globally fixed depth may still limit the integration of complementary
information across the visual hierarchy.

\subsection{RQ3. Why Does Factorization Matter?}
\label{sec:factorization}
We isolate three potential sources of improvement: access to
multi-depth information, learned cross-depth aggregation, and
factorized allocation. The repeated-layer controls preserve the full
constructor and its parameter count while removing cross-depth
complementarity. Uniform mean fusion retains all 24 depths without
learned allocation, whereas the parameter-matched monolithic variant
uses one shared allocation. We additionally test shared K/V projections and image-independent queries.

\begin{table}[t]
    \centering
    \caption{
        Controlled ablation of visual-target construction.
Repeated-layer variants preserve the complete constructor while
replacing all depth inputs with copies of one visual layer.
Uniform mean fusion directly averages all 24 per-layer embeddings
without the disentangled cross-attention module.
Results are Top-1 and Top-5 accuracy (\%).
    }
    \label{tab:neuroglyph_ablation}

    \scriptsize
    \setlength{\tabcolsep}{2.5pt}
    \renewcommand{\arraystretch}{1.05}

    \resizebox{\columnwidth}{!}{%
    \begin{tabular}{lcccccccc}
        \toprule
        \textbf{Variant}
        & \multicolumn{4}{c}{\textbf{EEG}}
        & \multicolumn{4}{c}{\textbf{MEG}} \\
        \cmidrule(lr){2-5}
        \cmidrule(lr){6-9}

        &
        \multicolumn{2}{c}{\textbf{Intra-subject}}
        &
        \multicolumn{2}{c}{\textbf{Inter-subject}}
        &
        \multicolumn{2}{c}{\textbf{Intra-subject}}
        &
        \multicolumn{2}{c}{\textbf{Inter-subject}} \\

        &
        \textbf{Top-1}
        & \textbf{Top-5}
        & \textbf{Top-1}
        & \textbf{Top-5}
        & \textbf{Top-1}
        & \textbf{Top-5}
        & \textbf{Top-1}
        & \textbf{Top-5} \\
        \midrule

        Repeated Final-Layer
        & 42.9 & 73.7
        & 11.1 & 32.1
        & 21.6 & 47.9
        & 2.9 & 11.3 \\

        Repeated Best-Layer
        & 79.2 & 95.9
        & 18.5 & 42.9
        & 35.0 & 66.0
        & 3.0 & 14.4 \\

        Monolithic-wide
        & 81.0 & 96.6
        & 21.2 & 48.4
        & 42.1 & 69.8
        & 3.9 & 13.9 \\

        Uniform Mean Fusion
        & 84.0 & 98.0
        & 20.2 & 46.9
        & 43.3 & 71.3
        & 3.8 & 14.1 \\

        w/ shared K/V
        & 84.4 & 97.7
        & 21.6 & 50.0
        & 44.8 & 72.4
        & 3.4 & 14.4 \\

        w/o image conditioning
        & 84.5 & 97.7
        & 20.4 & 48.8
        & 45.6 & 72.8
        & 3.5 & 14.3 \\

        \midrule

        \textbf{NeuroGlyph}
        & \textbf{84.8} & \textbf{98.1}
        & \textbf{22.4} & \textbf{51.2}
        & \textbf{46.0} & \textbf{73.3}
        & \textbf{4.0} & \textbf{14.8} \\

        \bottomrule
    \end{tabular}%
    }
\end{table}

As shown in Table~\ref{tab:neuroglyph_ablation}, the controlled
ablation separates the contribution of multi-depth information from
that of target-side capacity. Replicating the final-layer feature
across all depth inputs while preserving the full constructor reduces
Top-1 accuracy by 41.9, 11.3, 24.4, and 1.1 percentage points relative
to NeuroGlyph across the four settings. These results suggest that
target-side capacity alone does not explain the gains of NeuroGlyph and
cannot compensate for information missing from the final-layer
representation.

Uniform mean fusion further supports this interpretation. It outperforms
the repeated-best-layer variant by 4.8, 1.7, 8.3, and 0.8 Top-1 points,
showing that much of the gain comes from retaining the full visual
hierarchy rather than from additional parameters.

NeuroGlyph nevertheless improves over uniform mean fusion by 0.8, 2.2,
2.7, and 0.2 Top-1 points, with corresponding Top-5 gains of 0.1, 4.3,
2.0, and 0.7 points. Thus, equal averaging captures cross-depth
complementarity, but learned image-conditioned allocations exploit it
more effectively.

Despite having a matched parameter count, the monolithic variant
underperforms NeuroGlyph by 3.8, 1.2, 3.9, and 0.1 Top-1 percentage
points across the four settings, respectively, and does not
consistently surpass uniform mean fusion. These results suggest that
a single learned allocation over visual depths is insufficient and
support organizing depth allocation into multiple complementary
factors.

Finally, sharing K/V projections or removing image-conditioned queries
consistently reduces performance. Overall, the results show that
multi-depth information provides the main improvement over single-layer
targets, while factorized allocation, independent factor transformations,
and image conditioning provide further gains. NeuroGlyph achieves the
best result on all eight metrics.

Additional analyses of factor granularity and learned factor--depth profiles, including their cross-subject consistency, are provided in the supplementary material.

\subsection{RQ4. Does Gaussian Initialization Help?}
\label{app:init_ablation}
NeuroGlyph initializes the learnable factor--layer biases with
depth-wise Gaussian profiles, giving different factors distinct initial
depth preferences. To test whether the gains depend on this design, we
instead initialize all biases as $\beta^{(0)}_{k\ell}=0$, while keeping
them learnable and leaving other settings unchanged.

\begin{table}[htbp]
    \centering
    \caption{Effect of factor--layer bias initialization.
    Results are Top-1 and Top-5 accuracy (\%).}
    \label{tab:init_ablation}
    \resizebox{\linewidth}{!}{
    \begin{tabular}{lcccccccc}
        \toprule
        & \multicolumn{4}{c}{EEG}
        & \multicolumn{4}{c}{MEG} \\
        \cmidrule(lr){2-5}
        \cmidrule(lr){6-9}
        Initialization
        & \multicolumn{2}{c}{Intra}
        & \multicolumn{2}{c}{Inter}
        & \multicolumn{2}{c}{Intra}
        & \multicolumn{2}{c}{Inter} \\
        & Top-1 & Top-5 & Top-1 & Top-5
        & Top-1 & Top-5 & Top-1 & Top-5 \\
        \midrule
        Zero
        & 83.6 & 97.5 & 20.2 & 48.3
        & 45.6 & 72.0 & 3.4 & 14.1 \\
        \textbf{Gaussian}
        & \textbf{84.8} & \textbf{98.1}
        & \textbf{22.4} & \textbf{51.2}
        & \textbf{46.0} & \textbf{73.3}
        & \textbf{4.0} & \textbf{14.8} \\
        \bottomrule
    \end{tabular}}
\end{table}

Gaussian initialization consistently improves performance.
Zero initialization nevertheless remains competitive with the
parameter-matched monolithic constructor, outperforming it in
five of the eight metrics and showing particularly clear gains
in both intra-subject settings.

\section{Conclusion}
\label{sec:conclusion}

Brain-to-image retrieval typically uses the final visual layer as a fixed supervision target, overlooking information distributed across the visual hierarchy. A more suitable approach is to construct the target from multiple depths and adapt their contributions across target components and images. NeuroGlyph is proposed, which constructs a trial-independent target through factor-specific, image-conditioned allocation over multiple visual depths. NeuroGlyph outperforms final-layer supervision in all controlled comparisons and achieves the highest reported accuracy in six of eight system-level metrics across THINGS-EEG and THINGS-MEG.

Our analyses show that retrieval performance is highly sensitive to target depth, while no single fixed layer is optimal across modalities and evaluation protocols. Multi-depth information provides the main improvement, with factorized allocation and image conditioning offering further gains. These findings highlight the importance of visual-target construction in neural image retrieval.
\bibliography{aaai2027}


\clearpage

\twocolumn[
\begin{center}
    {\LARGE\bfseries Appendix\par}
    {\Large
    Supplemental Material for
    ``The Visual Target Matters: Learning across the Visual Hierarchy for Brain-to-Image Retrieval''
    \par}
\end{center}
]

\appendix

\section{Experimental Details}
\label{app:experimental_details}

\subsection{Datasets Protocols}
\label{app:datasets}

\paragraph{THINGS-EEG.}
THINGS-EEG contains recordings from 10 subjects collected over four
sessions.  The prescribed zero-shot split contains 1,654 training
concepts and 200 disjoint test concepts.  Each training concept is
represented by 10 images, each with four retained repetitions; we
average the repetitions of each image to obtain 16,540 training
examples per subject.  The test split contains one image per concept
with 80 repetitions, which are averaged to obtain 200 test queries per
subject.  We baseline-correct the epochs using the $-200$--$0$ ms
prestimulus interval, resample them to 250 Hz, and retain the
$0$--$1,000$ ms poststimulus window.  We apply session-wise multivariate
noise normalization, estimating the covariance from training epochs
only.  Following the implementation used by all compared target
constructors, we retain 17 posterior channels (P7, P5, P3, P1, Pz, P2,
P4, P6, P8, PO7, PO3, POz, PO4, PO8, O1, Oz, and O2), giving an input of
$17\times250$ samples per averaged response.

\paragraph{THINGS-MEG.}
THINGS-MEG contains recordings from four subjects over 12 sessions and
spans 1,854 THINGS concepts.  For the disjoint retrieval protocol, 200
concepts are reserved for testing and the remaining 1,654 concepts are
used for training.  The training split contains 12 distinct images per
concept with one response per image, yielding 19,848 training examples
per subject.  The test split contains one image for each of the 200
held-out concepts, repeated 12 times.  We use the provided preprocessed
epochs, retain all 271 MEG channels, crop each epoch to $0$--$1,000$ ms,
and represent it by $271\times201$ samples.  The 12 repeated test
responses are averaged image-wise before retrieval; averaging the
single-response training examples leaves them unchanged.

\subsection{Implementation Details}
\label{app:implementation}

\paragraph{Visual feature extraction.}
Unless otherwise stated, we use the frozen
 OpenGVLab/InternViT-300M-448px backbone at its native
$448\times448$ resolution and extract all 24 transformer-block outputs.
For each layer, we discard the class token and mean-pool the patch
tokens. The final pooled output is used to condition the factor queries;
if unavailable, it is obtained by mean-pooling the last-layer patch
tokens. All visual features are $\ell_2$-normalized, cached in FP32,
and kept frozen during training. Each layer feature is then processed by
an independent learned $1024\times1024$ projection, followed by
normalization, factor-specific decomposition, and factor-specific
LayerNorm before the key and value projections. The final pooled feature
is used only for query conditioning and is not concatenated into the
constructed target.

\paragraph{Neural encoders.}
Unless otherwise stated, we use EEGProject for both modalities.
It flattens the $17\times250$ EEG or $271\times201$ MEG input
and applies a linear projection to 1,024 dimensions, followed by
a residual block consisting of GELU, a second 1,024-dimensional
linear layer, dropout with rate 0.3, and LayerNorm. A learned
$1024\times1024$ retrieval projection maps this representation
to the common space, where it is $\ell_2$-normalized.
The EEG and MEG models use the same architecture but are trained
independently.

\paragraph{NeuroGlyph configuration.}
The default constructor uses $K=4$ factors of dimension $d_f=256$, so
the concatenated factor width is $Kd_f=1,024$; the final retrieval
dimension is also $d=1,024$.  Each visual depth has an independent
projection into the four factor channels, and each factor uses its own
key and value projections.  A learned base query is additively
conditioned on the normalized 1,024-dimensional final image feature.
We initialize the learnable factor--layer biases with the Gaussian depth
profiles in Equation~\eqref{eq:gaussian_bias_initialization}, whose
centers are distributed from shallow to deep layers.  These profiles are
an initialization rather than fixed assignments and carry no semantic or
biological labels.

\paragraph{Optimization.}
We jointly optimize the neural encoder and target constructor with
AdamW using a learning rate of $10^{-4}$, weight decay of $10^{-4}$,
and the default momentum parameters
$(\beta_1,\beta_2)=(0.9,0.999)$. The training batch size is 1,024,
while validation and test batches contain 200 examples per subject.
Training runs for at most 50 epochs with early stopping, using a
minimum improvement of $10^{-3}$ and a patience of five epochs. We
use neither learning-rate warmup nor a learning-rate scheduler. The
temperature of the symmetric contrastive loss is fixed at
$\tau=0.07$.

Experiments are conducted on a workstation
equipped with two NVIDIA RTX A6000 GPUs. Each individual run uses one
GPU, while the two GPUs are used to execute independent runs in
parallel. Training uses the default 32-bit PyTorch Lightning
precision. Visual features are extracted in batches of 64 and
cached before training. 

\section{Additional Experimental Results}
\label{app:additional_results}

\subsection{Generalization Across Neural Encoders}
\label{app:neural_encoder_generalization}

To examine whether NeuroGlyph depends on a particular neural encoder,
we evaluate five neural architectures while keeping the frozen visual
backbone, visual-target constructor, training objective, and evaluation
protocol unchanged. All experiments in this analysis use InternViT-300M-448px
as the frozen visual backbone.

As shown in Table~\ref{tab:appendix_neural_encoders}, NeuroGlyph is
compatible with all evaluated neural encoders, although their absolute
performance differs across datasets and evaluation protocols.
EEGProject achieves the strongest intra-subject performance on both
EEG and MEG, whereas TSConv obtains the strongest EEG inter-subject
result. TSConv and EEGProject achieve the same MEG inter-subject
Top-1 accuracy. These results indicate that NeuroGlyph is not tied to
a single neural architecture, while neural-encoder selection remains
important for absolute retrieval performance.

For consistency across the four main evaluation settings,
all main-text results use EEGProject as the default neural encoder.
Table~\ref{tab:appendix_neural_encoders} is an additional
encoder-generalization analysis and does not selectively replace
the default configuration reported in the main text.

\begin{table}[htbp]
    \centering
    \caption{
        Generalization of NeuroGlyph across neural encoders.
        All variants use the same frozen InternViT-300M-448px visual backbone, NeuroGlyph
        target constructor, training objective, and evaluation
        protocol. Only the neural encoder is changed.
        Results are Top-1 retrieval accuracy (\%).
    }
    \label{tab:appendix_neural_encoders}

    \scriptsize
    \setlength{\tabcolsep}{4.5pt}
    \renewcommand{\arraystretch}{1.08}

    \resizebox{\columnwidth}{!}{%
    \begin{tabular}{lcccc}
        \toprule
        \textbf{Neural encoder}
        & \textbf{EEG Intra}
        & \textbf{EEG Inter}
        & \textbf{MEG Intra}
        & \textbf{MEG Inter} \\
        \midrule

        DeepNet
        & 21.7
        & 12.8
        & 10.0
        & 3.4 \\

        ShallowNet
        & 46.5
        & 19.6
        & 17.3
        & 3.4 \\

        EEGNet
        & 57.8
        & 18.8
        & 19.5
        & 3.0 \\
        
        TSConv
        & 61.6
        & \textbf{25.4}
        & 24.9
        & \textbf{4.0} \\

        \midrule

        EEGProject
        & \textbf{84.8}
        & 22.4
        & \textbf{46.0}
        & \textbf{4.0} \\

        \bottomrule
    \end{tabular}%
    }
\end{table}

\subsection{Additional Ablation Studies}
\label{app:additional_ablations}

\paragraph{Sensitivity to factor granularity.}
We vary the number of NeuroGlyph factors
$K\in\{1,2,4,8,16\}$ and set $d_f=1024/K$, keeping the total factor
width $Kd_f$ fixed at 1024. The configuration
$(K,d_f)=(1,1024)$ is the single-factor version of NeuroGlyph,
using one 1024-dimensional key/value transformation branch and one
layer-allocation distribution. It differs from the parameter-matched
\emph{Monolithic-wide} baseline in Table~\ref{tab:neuroglyph_ablation},
which retains four independent 256-dimensional key/value
transformation branches, concatenates them before computing one shared
layer allocation, and averages their factor--layer bias vectors.
Thus, the two single-allocation variants use different internal
parameterizations and are not expected to produce identical results.

\begin{figure}[htbp]
    \centering
    \includegraphics[width=\linewidth]{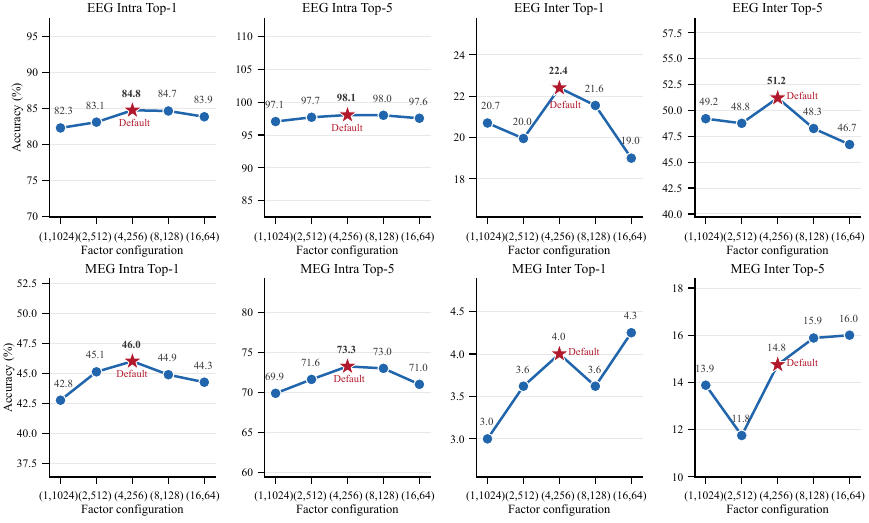}
    \caption{
        Sensitivity to factor granularity within NeuroGlyph under the
        fixed total width $Kd_f=1024$. The red star marks the default
        configuration $(K,d_f)=(4,256)$. The $(1,1024)$ point denotes
        single-factor NeuroGlyph rather than the Monolithic-wide
        baseline in Table~\ref{tab:neuroglyph_ablation}.
    }
    \label{fig:analysis3}
\end{figure}

Performance does not increase monotonically with $K$. The
configuration $(K,d_f)=(4,256)$ performs best in six of the
eight metrics and provides the strongest overall trade-off.
The finer configuration $(16,64)$ performs best only for the
two MEG inter-subject metrics, exceeding the default by
0.3 percentage points in Top-1 and 1.2 points in Top-5.
We therefore use $(4,256)$ as the default configuration in all
main-text experiments, rather than selecting $K$ separately for
each evaluation setting.

\section{Additional Analysis}
\label{app:additional_analysis}

\subsection{Analysis of Learned Factor--Depth Allocations}
\label{sec:learned_organization}
\label{app:allocation_analysis}

\begin{figure}[htbp]
    \centering
    \includegraphics[width=\linewidth]{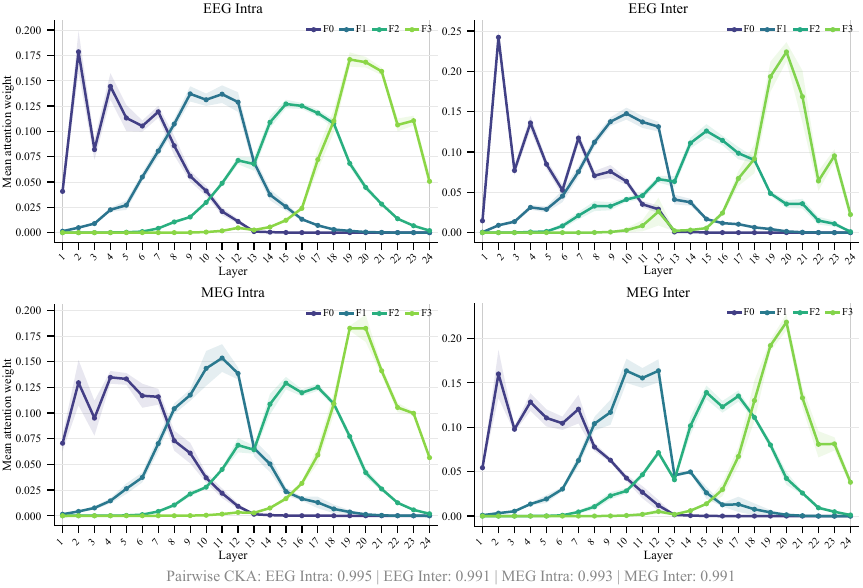}
    \caption{
        Mean factor--layer allocation profiles with cross-subject
        standard deviations. Pairwise CKA measures the similarity
        between subject-specific allocation matrices.
    }
    \label{fig:analysis4}
\end{figure}

\paragraph{Factor--depth preferences and cross-subject consistency.}
As shown in Figure~\ref{fig:analysis4}, the learned factor--layer
allocation profiles exhibit distinct depth preferences. Each curve is averaged across subjects, and the
shaded region denotes the corresponding cross-subject standard
deviation.

Across all four settings, the factors exhibit distinct and partially
overlapping depth preferences. Factor $F_0$ concentrates on shallow
representations, $F_1$ on early-to-middle depths, $F_2$ on
middle-to-late depths, and $F_3$ on deeper representations. The
factors therefore do not collapse onto one common layer or
concentrate exclusively on the visual endpoint.

The mean pairwise CKA between subject-specific allocation matrices
ranges from 0.991 to 0.995, indicating highly consistent
factor--depth profiles across subjects. Because the factor--layer
biases are initialized with ordered depth preferences, this analysis
demonstrates non-collapse and cross-subject consistency, but does not
establish that the ordering emerges independently of initialization.
It also does not imply semantic disentanglement, statistical
independence, functional complementarity, or biological
correspondence among the factors.

\subsection{Computational Cost}
\label{app:computational_cost}

\paragraph{Trainable parameter count.}
We analyze the trainable parameters introduced by the NeuroGlyph
visual target constructor. The frozen visual backbone is excluded
because it is not updated during neural--visual alignment.

The default configuration uses all $L=24$ visual depths of InternViT-300M-448px,
$K=4$ target-construction factors, a factor dimension of
$d_f=256$, and a retrieval dimension of $d=1024$. Therefore,
the total factor width satisfies $Kd_f=d=1024$. The visual
representations produced by the selected backbone also have dimension
$d_{\mathrm{vis}}=1024$.

\begin{table}[htbp]
    \centering
    \caption{
        Trainable parameter breakdown of the NeuroGlyph visual target
        constructor under the default configuration
        ($L=24$, $K=4$, $d_f=256$, and
        $d_{\mathrm{vis}}=d=1024$).
        The frozen visual backbone and neural-side modules are excluded.
    }
    \label{tab:target_constructor_parameters}

    \scriptsize
    \setlength{\tabcolsep}{3.2pt}
    \renewcommand{\arraystretch}{1.08}

    \resizebox{\columnwidth}{!}{%
    \begin{tabular}{lccc}
        \toprule
        \textbf{Component}
        & \textbf{Parameterization}
        & \textbf{Parameters}
        & \textbf{Share} \\
        \midrule

        Layer-wise visual projections
        & $L(d_{\mathrm{vis}}d+d)$
        & $25{,}190{,}400$
        & $47.5\%$ \\

        Per-layer factor decomposition
        & $L\!\left[d(Kd_f)+Kd_f\right]$
        & $25{,}190{,}400$
        & $47.5\%$ \\

        Factor-wise allocation and fusion
        & Independent of $L$
        & $2{,}624{,}608$
        & $5.0\%$ \\

        \midrule

        \textbf{Target-constructor total}
        & --
        & $\mathbf{53{,}005{,}408}$
        & $\mathbf{100.0\%}$ \\

        \bottomrule
    \end{tabular}%
    }
\end{table}

The NeuroGlyph visual target constructor contains approximately
$53.01$M trainable parameters. Most of these parameters arise from
the two layer-specific transformations. The layer-wise visual
projections and per-layer factor decomposition each contain
approximately $25.19$M parameters and jointly account for $95.0\%$
of the target constructor. Both components scale linearly with the
number of selected visual depths $L$.

In contrast, the factor-wise depth-allocation and output-fusion
modules contain only $2.62$M parameters, corresponding to $5.0\%$
of the target constructor. Thus, the main parameter cost of
NeuroGlyph arises from independently transforming representations
at different visual depths rather than from the factor-specific
allocation mechanism itself.

\paragraph{Offline target construction and online retrieval.}
The visual backbone remains frozen throughout training, allowing its
hierarchical representations to be extracted and cached in advance.
At inference time, the NeuroGlyph constructor receives only image
features. The target of each gallery image can therefore be computed
once and stored in a target bank.

Given a new neural recording, online retrieval requires only one
forward pass through the neural encoder followed by cosine-similarity
ranking against the precomputed image targets. For a gallery containing
$N_g$ images, the ranking complexity is
$\mathcal{O}(N_g d)$, which is identical to conventional
single-embedding retrieval with the same retrieval dimension.



\end{document}